\documentclass[journal]{IEEEtran}
\usepackage{amssymb}
\usepackage{color}
\usepackage{multirow}
\usepackage{mathrsfs}
\usepackage{flushend}

\newtheorem{definition}{Definition}

\usepackage{cite}
\ifCLASSINFOpdf
\usepackage[pdftex]{graphicx}
\graphicspath{{../pdf/}{../jpeg/}}
\DeclareGraphicsExtensions{.pdf,.jpeg,.png}
\else
\usepackage[dvips]{graphicx}
\graphicspath{{../eps/}}
\DeclareGraphicsExtensions{.eps}
\fi
\usepackage[cmex10]{amsmath}
\usepackage{array}
\usepackage[tight,footnotesize]{subfigure}
\usepackage{stfloats}

\usepackage{algpseudocode}
\usepackage{algorithm}
\usepackage{setspace}
\usepackage{arydshln}
\usepackage{rotating}
\usepackage{microtype}
\IEEEoverridecommandlockouts
\usepackage{epstopdf}

\begin{document}
\title{Towards a Robust Classifier: An MDL-Based Method for Generating Adversarial Examples}

\author{\IEEEauthorblockN{Behzad Asadi and Vijay Varadharajan}\\
\IEEEauthorblockA{The University of Newcastle, Australia}\vspace{-20pt}}

\maketitle
\begin{abstract}
We address the problem of adversarial examples in machine learning where an adversary tries to misguide a classifier by making functionality-preserving modifications to original samples. We assume a black-box scenario where the adversary has access to only the feature set, and the final hard-decision output of the classifier. We propose a method to generate adversarial examples using the minimum description length (MDL) principle. Our final aim is to improve the robustness of the classifier by considering generated examples in rebuilding the classifier. We evaluate our method for the application of static malware detection in portable executable (PE) files. We consider API calls of PE files as their distinguishing features where the feature vector is a binary vector representing the presence-absence of API calls.  In our method, we first create a dataset of benign samples by querying the target classifier. We next construct a code table of frequent patterns for the compression of this dataset using the MDL principle. We finally generate an adversarial example corresponding to a malware sample by selecting and adding a pattern from the benign code table to the malware sample. The selected pattern is the one that minimizes the length of the compressed adversarial example given the code table. This modification preserves the functionalities of the original malware sample as all original API calls are kept, and only some new API calls are added. Considering a neural network, we show that the evasion rate is 78.24 percent for adversarial examples compared to 8.16 percent for original malware samples. This shows the effectiveness of our method in generating examples that need to be considered in rebuilding the classifier.

\end{abstract}
\begin{IEEEkeywords}
Classification, Minimum Description Length, Adversarial Examples, Malware Detection, Transactional Dataset
\end{IEEEkeywords}	
\IEEEpeerreviewmaketitle
\section{Introduction}
Machine learning algorithms have been widely deployed in different applications as automated decision-making tools due to their generalization capabilities. This includes autonomous driving, and cybersecurity applications where false decisions can have serious consequences. However, machine learning algorithms have shown to be vulnerable in adversarial settings. An adversary can influence the decisions of a machine learning model both in the training phase and in the test phase.

In this work, we address the problem of adversarial examples in the test phase. This is where an adversary modifies original samples to misguide a trained classifier. Modifications cannot be arbitrary, and must preserve the functionalities of original samples. Most of the existing works addressing this problem have been in the area of image classification. In image classification, adversarial examples are generated by adding perturbations to normal images; perturbations must be imperceptible to human eyes to be considered as functionality-preserving. There has also been some works in the area of malware detection which is also our considered application in this work. In malware detection, adversarial examples are generated by modifying malware samples to be misclassified as benign samples; modifications need to satisfy a set of constraints defined to preserve the functionalities of original malware samples.

Adversarial examples can be classified based on the adversary specificity into targeted or non-targeted examples~\cite{Survey1}. Targeted examples are generated to be misclassified into a specific wrong class. Non-targeted examples are generated to be misclassified into an arbitrary wrong class. Adversarial examples can also be classified based on the adversary knowledge of the classifier into white-box, and black-box examples~\cite{Survey1}. White-box examples are generated by an adversary which has access to all the parameters of the target trained classifier~\cite{FGSMSingleStep, FGSMMultiStep,JSMA, DeepFool, FGSMMalware, JSMAMalware}. In white-box scenarios, adversarial examples are generated using gradient-based methods where the gradient can be computed by knowing all the parameters of the classifier. Black-box examples are generated by an adversary which does not have access to the parameters of the target trained classifier; the adversary has access to only the output of the classifier in the form of either soft decision (confidence score)~\cite{BalckBoxImageConfidenceScore,BalckBoxMalwareConfidenceScore} or hard decision (label)~\cite{BalckBoxImageJSMA, BalckBoxImageGAN1, BalckBoxImageGAN2, BalckBoxMalwareGAN1, BalckBoxMalwareGAN2, BalckBoxMalwareAPICall, BalckBoxMalwareNoFeature}. In black-box scenarios where the adversary has access to the hard-decision output of the classifier, adversarial examples are mainly generated by first constructing a substitute model for the target model, and then using white-box methods for the substitute model. This idea is based on the transferability of adversarial examples~\cite{Transferability} which says that adversarial examples constructed to misguide a specific classifier can also misguide another classifier with a totally different architecture.

As the final objective, we are trying to improve the robustness of machine learning algorithms by taking into account adversarial examples. There are some existing defense mechanisms to make a classifier more robust to adversarial examples such as defense distillation~\cite{DistillationDefense}, and adversarial training~\cite{AdversarialTraining0,AdversarialTraining1,AdversarialTraining2}. Distillation helps the classifier to generalize better to slightly modified samples, and consequently becomes more robust to adversarial examples. In adversarial training, adversarial examples are generated and utilized during the retraining process. Both white-box and black-box examples are considered in the training process. Therefore, the classifier gets exposed to such examples during the training process.

\subsection{Existing Works and Contributions}
In this work, we address the problem of adversarial examples in a black-box scenario where the adversary has access to the feature space and the hard-decision output of the target classifier. We propose an approach to generate adversarial examples using the minimum description length (MDL) principle.

We consider the application of static malware detection in portable executable (PE) files where their API calls are used to decide whether they are benign or malicious. As mentioned earlier, in malware detection, adversarial examples are generated by making functionality-preserving modifications to original malware samples such that they are misclassified as benign samples. Hu and Tan~\cite{BalckBoxMalwareGAN2} addressed this application using a dataset consisting of 160 different API calls. Their approach is based on constructing a substitute model for the target classifier and using generative adversarial networks~\cite{GAN}. We address this application using a dataset consisting of a much larger number of features; our dataset consists of 22761 unique API~calls. 

In our MDL-based approach, we first create a dataset of samples all identified as benign samples by querying the target classifier. We then construct a code table of frequent patterns for the compression of samples in this benign dataset using the MDL principle. We finally generate an adversarial example corresponding to a malware sample by selecting and adding a pattern from the benign code table to the malware sample. The selected pattern is the one that minimizes the length of the compressed adversarial example given the benign code table. Note that, in our method, we do not construct a substitute model as we only need a dataset of benign samples. Also, our method preserves the functionalities of malware samples as only some new API calls are added to malware samples without removing any existing ones. Considering a neural network as the classifier, using our method, the evasion rate for adversarial examples is 78.24 percent compared to 8.16 percent for original malware samples. This shows the necessity for considering these generated adversarial examples in rebuilding the neural network.

\section{Static Malware Detection}\label{Sec:SMD}
Malware detection is one of the areas that machine learning algorithms have been able to contribute. Traditional algorithms for malware detection search for known patterns which requires them to have a copy of all malware samples. These algorithms are not effective nowadays as (i) polymorphism is used within a malware family, (ii) the number of new malware families is increasingly growing, and (iii) they are not capable of zero-day malware detection. This makes machine learning algorithms good candidates for automated malware detection. This is because they can extract complex patterns using different attributes of a malware, and they can also help with zero-day malware detection as they can generalize to new samples~\cite{MLforMalware}.

Malware detection can be divided into two main categories of dynamic (behavioral) and static (code) malware detection. In dynamic malware detection, samples are executed, and their run-time behavior is monitored to create indicators of malicious activities. In static malware detection, binary codes of samples are examined without executing them to create indicators of malicious activities.

As mentioned earlier, we consider static malware detection in PE files. Different types of features have been used for this task such as API calls~\cite{FGSMMalware}, byte-level N-grams~\cite{StaticNgram}, features from the PE header~\cite{StaticHeader}, and a combination of different types of features~\cite{EMBER}. We consider API calls of PE files to distinguish between malware and benign samples. The presence-absence of API calls forms a transactional dataset which is explained in the following section.

\section{Transactional Datasets}\label{Sec:TransactionalDataset} 
In this section, we present some preliminaries for transactional datasets required in this work. A transactional dataset, denoted by $\mathcal{D}$, is a non-empty multiset (bag) of transactions, i.e., $\mathcal{D}=\{\mathcal{T}_1,\mathcal{T}_2,\ldots,\mathcal{T}_n\}$. Each transaction is a subset of $\mathcal{I}=\left\{1,2,3,\ldots,m\right\}$ where $\mathcal{I}$ represents the set of all items (i.e., $\mathcal{T}_j\subseteq \mathcal{I}\;\forall j$).  We say that a transaction $\mathcal{T}_j$ supports an itemset $\mathcal{P}$ (which is also a subset of $\mathcal{I}$) if $\mathcal{P}\subseteq \mathcal{T}_j$. The support of an itemset $\mathcal{P}$, denoted by $sup(\mathcal{P})$, is the number of transactions that support the itemset. Considering that $\mathcal{D}(\mathcal{P})$ is the multiset of transactions that support the itemset $\mathcal{P}$, and $\mathcal{D}(\mathcal{Q})$ is the multiset of transactions that support the itemset $\mathcal{Q}$, we therefore have

\begin{itemize}
	\item $sup(\mathcal{P})=\left|\mathcal{D}(\mathcal{P})\right|$,
	\item $\mathcal{D}(\mathcal{P}\cup \mathcal{Q})=\mathcal{D}(\mathcal{P})\cap \mathcal{D}(\mathcal{Q})$,
	\item If $\mathcal{P}\subseteq\mathcal{Q}$, then $\mathcal{D}(\mathcal{P})\supseteq\mathcal{D}(\mathcal{Q})$,
	\item If $\mathcal{P}\subseteq\mathcal{Q}$, then $sup(\mathcal{P})\geq sup(\mathcal{Q})$,
\end{itemize}
where $|\cdot|$ denotes the cardinality of the multi-set. An itemset is considered to be \textit{frequent} if its support is greater than or equal to a user-decided threshold, denoted by $minsup$. A frequent itemset is \textit{closed} if it has no superset with the same support.

\section {MDL Principle and Its Applications}\label{Sec:mdl}
In this section, we present the MDL principle, and its applications for classification, and pattern summarization.
\subsection {MDL Principle}
Kolmogorov complexity theory, also known as algorithmic information theory, was developed to measure the information in objects in isolation, i.e., without knowing the distribution underlying the object. As in data mining, we normally do not know the underlying distribution of our data, we use algorithmic information theory to measure the information in our data. The Kolmogorov complexity of an object is the descriptive complexity of that object which is the length of the shortest computer program that can describe the object. This is formally defined as follows~\cite{ITBook}. 

\begin{definition}
	The Kolmogorov complexity of an object $x$ with respect to a universal computer $\mathcal{U}$, denoted by  $K_{\mathcal{U}}(x)$, is defined as
	\begin{align*}
	K_{\mathcal{U}}(x)=\underset{prog:\mathcal{U}(prog)=x}{\min} \ell(prog),
	\end{align*}
	which is the minimum length over all programs that print $x$ and halt.
\end{definition}

However, we cannot compute the Kolmogorov complexity of an object. Therefore, in practice, the MDL principle is utilised. Using the crude MDL version, we choose a model from a set of models, $\mathcal{M}$, that minimises the two-term objective function $\ell(x\mid M_i)+\ell(M_i)$ where $\ell(x\mid M_i)$ is the number of bits required to describe the object given the model, and $\ell(M_i)$ is the number of bits required to describe the model itself. Hence, based on the crude MDL, we have
\begin{align*}
\ell_\text{best}(x)=\underset{M_i\in\mathcal{M}}{\min}\; (\ell(x\mid M_i)+\ell(M_i)).
\end{align*}


\subsection{MDL-based Classifier}\label{Sec:MDLClassifier}
We here explain how to utilize the MDL principle to build a binary classifier. Supervised learning consists of two phases of training and test. In the training phase, we select a model for the training dataset of each class based on the MDL criterion,
\begin{align*}
M_{\mathcal{D}_1}&=\underset{M_i\in\mathcal{M}}{\arg\min}\; (\ell(\mathcal{D}_1\mid M_i)+\ell(M_i)),\\
M_{\mathcal{D}_2}&=\underset{M_i\in\mathcal{M}}{\arg\min}\; (\ell(\mathcal{D}_2\mid M_i)+\ell(M_i)).
\end{align*}

In the test phase, if for the transaction $\mathcal{T}$, we have   
\begin{align*}
\ell(\mathcal{T}\mid M_{\mathcal{D}_2})\leq\ell(\mathcal{T}\mid M_{\mathcal{D}_1}),
\end{align*}
this implies that
\begin{align*}
Pr(\mathcal{T}\mid \mathcal{D}_2)\geq Pr(\mathcal{T}\mid \mathcal{D}_1).
\end{align*}
Consequently, we classify the sample $\mathcal{T}$ under the second class. Otherwise, we classify it under the first class. Note that the term $\ell(M)$ in the crude MDL criterion prevents the model to be overfitted during the training phase. This is because, by using a complex (overfitted) model, we can minimize the term $\ell(\mathcal{D}\mid M)$. Therefore, using only this term as the selection criterion can result in overfitting. By considering both $\ell(\mathcal{D}\mid M)$ and $\ell (M)$ terms in the selection criterion, this scenario can be avoided.

\subsection{MDL-based Pattern Summarization}\label{Sec:CodeTable}

The MDL principle can be used for pattern summarization where we want to select a small subset of an existing large set of candidate patterns denoted by $\mathcal{F}$. In this part, we present the algorithm proposed by Vreeken et al.~\cite{KRIMP} which uses the MDL principle for pattern summarization. This algorithm performs pattern summarization by searching among code tables of patterns as the family of models to describe the data. A code table, denoted by $CT$, has two columns: the first column consists of selected patterns, and the second column consists of binary codes used to encode the patterns in the first column. This algorithm, which basically outputs a semi-adaptive  compression dictionary, selects the best code table as
\begin{align}\label{mdlct}
CT_{\text{best}}=\underset{CT}{\arg\min}\; (\ell(\mathcal{D}\mid CT)+\ell(CT)).
\end{align}

In the algorithm proposed by Vreeken et al.~\cite{KRIMP}, as the search space for constructing code tables is very large, a heuristic approach is used to select the best code table. This heuristic approach consists of three steps. In the first step, candidate patterns in the set $\mathcal{F}$ are ordered descending first by their support, second by their length. In the second step, a standard code table consisting of all singleton items is constructed. In the third step, candidate patterns from the ordered $\mathcal{F}$ are examined one by one. In this step, if adding a candidate pattern to the current code table results in a smaller objective function, i.e., $\ell(\mathcal{D}\mid CT)+\ell(CT)$, it is kept in the code table, otherwise it is dropped. This leads to keeping only a small subset of $\mathcal{F}$ in the final code table. The final code table is considered as the selected model by the MDL principle, and the patterns in the final code table are considered as the patterns chosen by the MDL principle.

We here explain how the two terms $\ell(\mathcal{D}\hskip-2pt\mid\hskip-2pt CT)$ and $\ell(CT)$ in equation~\eqref{mdlct} are calculated. The first term in equation~\eqref{mdlct}, $\ell(\mathcal{D}\mid CT)$, is calculated as
\begin{align*}
\ell(\mathcal{D}\mid CT)=\sum_{\mathcal{T}\in\mathcal{D}}\ell(\mathcal{T}\mid CT)=\sum_{\mathcal{T}\in\mathcal{D}}\sum_{\mathcal{P}\in \mathcal{C}(\mathcal{T})}\ell(\mathcal{P}\mid CT),
\end{align*}
where $\ell(\mathcal{P}\mid CT)$ is the length of the binary code for the pattern $\mathcal{P}$ in the second column, and $\mathcal{C}(\mathcal{T})$ is the set of patterns used to cover $\mathcal{T}$. The patterns covering a transaction satisfy the following properties
\begin{align*}
\forall\;\mathcal{P}_i,\mathcal{P}_j\in \mathcal{C}(\mathcal{T}),\text{ if } \mathcal{P}_i\neq\mathcal{P}_j \text{ then } \mathcal{P}_i\cap\mathcal{P}_j=\emptyset,
\end{align*}
and 
\begin{align*}
\bigcup_{\mathcal{P}\in \mathcal{C}(\mathcal{T})}\mathcal{P}=\mathcal{T}.
\end{align*}
As there can be several ways (different sets of patterns) to cover a transaction, the patterns in the code table are ordered descending first by their length, next by their support; the patterns are selected according to this order to cover a transaction.  

The lengths of binary codes in the second column of the code table, i.e., $\ell(\mathcal{P}\mid CT)$, are determined by the Shannon code which is a prefix code. The more a pattern used in the cover of transactions, the shorter its code. Therefore, by defining the usage of a pattern $\mathcal{P}$ as
\begin{align*}
\mathit{usage}(\mathcal{P})=\left|\left\{\mathcal{T}\in\mathcal{D}:\mathcal{P}\subseteq \mathcal{C}(\mathcal{T})\right\}\right|, 
\end{align*}
the code for the pattern $\mathcal{P}$ is of length
\begin{align*}
\ell(\mathcal{P}\mid CT)&=\left \lceil  -\log Pr(\mathcal{P}\mid D)\right \rceil\\
&=\left \lceil-\log \left(\frac{usage(\mathcal{P})}{\sum_{\mathcal{P'}\in CT}usage(\mathcal{P'})}\right)\right \rceil.
\end{align*}

The second term in equation~\eqref{mdlct}, $\ell(CT)$, is calculated as
\begin{align}\label{lengthcodetabel}
\ell(CT)&= \sum_{i\in\mathcal{I}} n_i\log (|\mathcal{I}|+1)+\sum_{\mathcal{P}\in CT}\log (|\mathcal{I}|+1)\nonumber\\&\hskip12pt+\sum_{\mathcal{P}\in CT}\ell(\mathcal{P}\mid CT),
\end{align}
where $n_i$ is the number of times that item $i$ appears in the patterns in the first column of the code table. The number of all possible items in first column of the code table considering a separator between each two patterns is $|\mathcal{I}|+1$. The first two terms on the left-hand side of equation~\eqref{lengthcodetabel} correspond to encoding the first column of the code table. The last term on the left-hand side of equation~\eqref{lengthcodetabel} corresponds to encoding the second column of the code table consisting of prefix binary codes.

\subsubsection{Example}
We here provide an example for pattern summarization. In this example, we consider the following dataset which consists of five items and 10 transactions. 
\begin{center}
	\begin{tabular}{ccccc}
		1&  2&  3& 4 & 5\\
		\hline
		1&  1&  1& 1& 0\\
		1&  1&  1& 1& 0\\ 
		1&  1&  0& 1& 0\\  
		0&  1&  1& 1& 1\\ 
		0&  0&  1& 1& 1\\
		0&  0&  0& 1& 1\\ 
		0&  1&  0& 0& 0\\
		0&  0&  1& 0& 0\\
		0&  0&  0& 1& 0\\
		0&  0&  0& 0& 1
	\end{tabular}
\end{center}
~\\
Each row represents a transaction. This dataset can be represented as
\begin{align*}
\mathcal{D}&=\left\{\{1,2,3,4\}^2,\{1,2,4\},\{2,3,4,5\},\right.\\
&\hskip90pt\left.\{3,4,5\},\{4,5\},\{2\},\{3\},\{4\},\{5\}\right\},
\end{align*}
where $\mathcal{D}$ is a multiset, and the superscript for an element shows the multiplicity of that element. We perform closed frequent pattern mining (CFPM) with $minsup=1$ to extract all closed frequent patterns (CFPs) of this dataset. This is to form the list of candidate patterns required to construct an MDL-based code table for this dataset. In this work, we use the Linear Time Closed Itemset Mining (LCM) algorithm~\cite{LCM} for CFPM. Using extracted CFPs, the ordered list of candidate patterns is

\begin{center}
	\begin{tabular}{c|c}
		$\mathcal{P}$&  $sup(\mathcal{P})$  \\
		\hline
		$\{4\}$&  7  \\ 
		$\{3\}$&  5  \\  	
		$\{2\}$&  5\\
		$\{3,4\}$&  4  \\  	
		$\{2,4\}$&  4\\
		$\{5\}$&  4  \\ 
		$\{2,3,4\}$&  3  \\  	
		$\{1,2,4\}$&  3\\
		$\{4,5\}$&  3  \\ 
		$\{1,2,3,4\}$&  2  \\  	
		$\{3,4,5\}$&  2  \\ 
		$\{2,3,4,5\}$&  1
	\end{tabular}
\end{center}
and the final code table using the described approach is

\begin{center}
	\begin{tabular}{|c|c|}
		\hline 
		$\mathcal{P}$ & binary code length\\  
		\hline 
		$\left\{1,2,4\right\}$   &  $3$  \\ 
		\hline 
		$\left\{4\right\}$     &  $3$  \\ 
		\hline 
		$\left\{3\right\}$     &  $2$  \\
		\hline 
		$\left\{2\right\}$     &  $4$  \\	
		\hline
		$\left\{5\right\}$     &  $3$  \\	
		\hline
		$\left\{1\right\}$     &  $8$  \\	
		\hline
	\end{tabular} 
\end{center}
~\\
This shows the effectiveness of the MDL principle for pattern summarization. In the second column of the code table, we have provided the lengths of binary codes than binary codes themselves. This is because the lengths are important than the codes themselves. Note that item 1 does not appear in the cover of any transactions, i.e., its usage is equal to zero. We keep all singleton items in the final code table by giving them a small usage when their usage is zero. This is to be able to cover any unseen transactions. 

\begin{figure*}[t]
	\centering
	\includegraphics[width=0.85\textwidth]{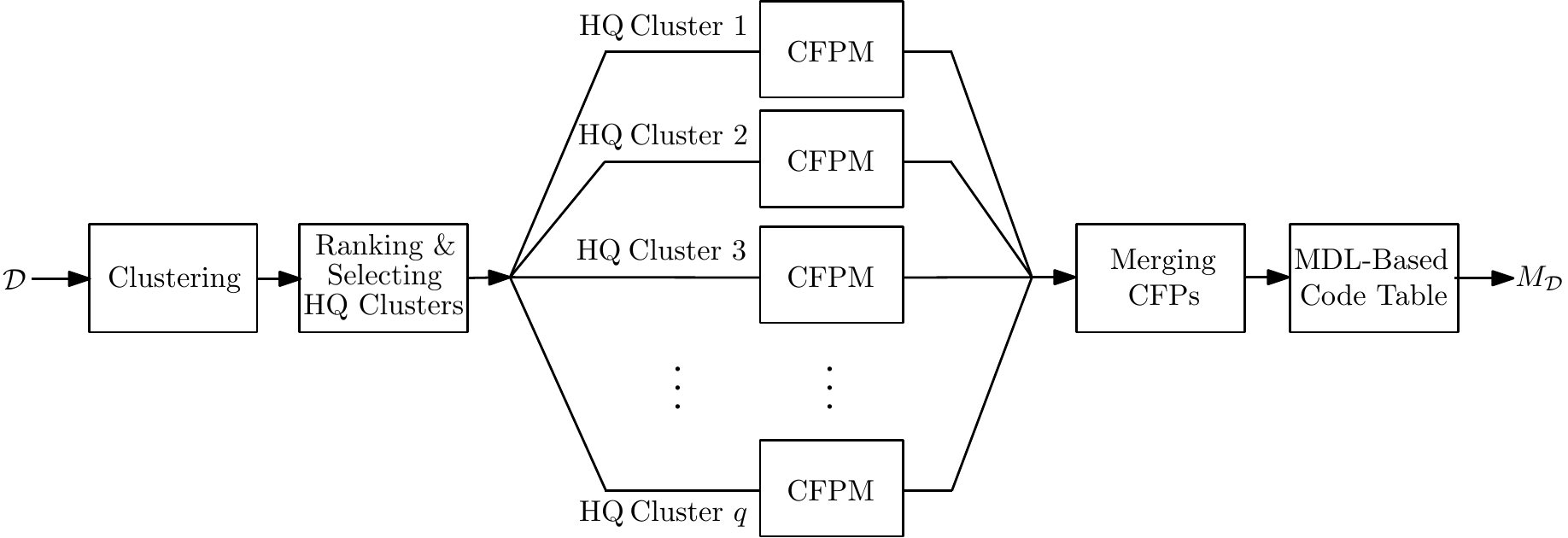}
	\caption{MDL-based model selection Method.} 
	\label{Fig:ProposedMethod}
\end{figure*}

\section{MDL-Based Model Selection}\label{OurMDLModel}
In this section, we explain our recently proposed MDL-based model-selection method~\cite{OurMDLClassifier}, shown in Fig.~\ref{Fig:ProposedMethod}. The method used for the example in Section~\ref{Sec:CodeTable} can be computationally very expensive for large datasets. This is because we may face pattern explosion in extracting all CFPs. To address this problem, we have recently proposed a method where we use clustering in conjunction with CFPM to form the list of candidate patterns. We have shown that this approach extracts a subset of all CFPs by giving priority to longer CFPs. This is important as the compression is mainly achieved through longer patterns. In our method, we first cluster the dataset. We use the Clustering with sLOPE (CLOPE) algorithm~\cite{clope} which is a fast algorithm for clustering transactional datasets. In the CLOPE algorithm, we do not need to know the number of clusters in advance, but we need to set the maximum number of clusters. This can be decided based on the parameter $minsup$. The larger the parameter $minsup$, the smaller the maximum number of clusters. For a large $minsup$, we do not face pattern explosion, and therefore we do not need clustering. After clustering, we rank clusters according to the following criterion
\begin{equation}\label{clusterquality}
Quality(\mathbb{C}_i)=\frac{H(\mathbb{C}_i)}{\left|\mathbb{C}_i\right|},
\end{equation}
where $H(\mathbb{C}_i)$ and $\left|\mathbb{C}_i\right|$ are the height and the number of transactions of cluster~$i$ respectively. The height of cluster $\mathbb{C}_i$ is defined as
\begin{equation*}
H(\mathbb{C}_i)=\frac{\sum_{\mathcal{T}_j\in \mathbb{C}_i}|\mathcal{T}_j|}{|\cup_{\mathcal{T}_j\in \mathbb{C}_i}\mathcal{T}_j|}.
\end{equation*}
The cluster quality takes a value between zero and one. It is equal to one where all the transactions of a cluster are the same (the highest quality). We next select a subgroup of clusters as high-quality (HQ) clusters by setting a quality threshold, and perform CFPM in only HQ clusters. In HQ clusters, transactions share majority of their items, and as a result the number of CFPs in these clusters is not large even by considering a small $minsup$. Low-quality (LQ) clusters are the main reason for pattern explosion, and the output of CFPM in these clusters consists of mainly short patterns.

As the output of the pattern-mining stage, we take the union over the outputs of CFPM in HQ clusters. We finally construct a code table of patterns according to Section~\ref{Sec:CodeTable} as the selected model.

\section{Proposed MDL-Based Adversarial Examples}\label{Sec:proposedmethod}
In this section, we propose an MDL-based method to generate adversarial examples. In our recent work~\cite{OurMDLClassifier}, we designed an MDL-based classifier where we showed that one of the main advantages of using the MDL principle for the task of classification is about interpretability. This means that, to some extent, we can explain the reasons why a sample is classified under a specific class rather than other classes. As we saw in Section~\ref{Sec:MDLClassifier}, the better a sample is compressed using the model of a class, the higher the probability that the sample belongs to that class. Therefore interpretability is about finding the reasons why a sample can be compressed better using the model of a specific class. This can be done by considering the structure of the family of the models chosen for compression. This motivates utilizing the MDL principle to define a metric to generate adversarial examples. By knowing the reasons why a sample can be compressed better using a model, it is possible to modify the sample to have a shorter compressed version considering the model for a wrong class. This increases the probability that the sample is classified under the wrong class.

We consider a black-box scenario where the adversary has access to only the input and the hard-decision output of the target classifier. In our method, we first select a class, and construct a dataset $\mathcal{D}_\text{sel}$ in which all samples are classified under the selected class by the target classifier. This is done by querying the target classifier. We then choose a model that best describes the dataset $\mathcal{D}_\text{sel}$ using the MDL principle,
\begin{align}\label{Eq: ModelSelection}
M_{\mathcal{D}_\text{sel}}=\underset{M_i\in\mathcal{M}}{\arg\min}\; (\ell(\mathcal{D}_\text{sel}\mid M_i)+\ell(M_i)).
\end{align}
Now considering that $\mathcal{T}$ is a sample not belonging to the selected class, we generate an adversarial example $\mathcal{T}_\text{adv}$ corresponding to $\mathcal{T}$ based on the following metric
\begin{align}\label{Eq: MDLAdversarial}
\mathcal{T}_{\text{adv}}= \underset{\mathcal{T}'\in\mathcal{S}(\mathcal{T})}{\arg\min}\;\ell(\mathcal{T}'\mid M_{\mathcal{D}_\text{sel}}),
\end{align}
where $\mathcal{S}(\mathcal{T})$ represents the set of all modified versions of $\mathcal{T}$ with the same functionalities. In our approach, we are actually trying to choose the vector $\mathcal{T}'$ from $\mathcal{S}(\mathcal{T})$ which has the maximum $P(\mathcal{T}'\mid \mathcal{D}_\text{sel})$. This is to misguide the classifier to identify the adversarial example $\mathcal{T}_{\text{adv}}$ as a member of the selected class.

In the application of malware detection, adversarial examples are generated by modifying malwares samples to be misclassified as benign samples. Therefore, we first need to construct a dataset of samples all identified as benign by the target classifier. We then, using~\eqref{Eq: ModelSelection}, choose a model that best describes the benign dataset. We finally, using~\eqref{Eq: MDLAdversarial}, generate adversarial examples corresponding to original malware samples.  

Note that in our proposed approach, we do not construct a substitute model for the original classifier. This is as opposed to black-box methods in which, first, a substitute model is constructed, and then, a white-box method is employed to generate adversarial examples for the substitute model. For instance, in the application of malware detection, for creating a substitute model, we need a dataset of malware samples in addition to a dataset of benign samples. In our approach, we only need a dataset of benign samples which is easier to construct than a dataset of malware samples.

\subsection{Algorithm}\label{Sec:proposedalg}
We here propose an algorithm as a suboptimal implementation of our approach for the application of static malware detection in PE files. API calls of PE files are used as distinguishing features. As mentioned in Section~\ref{Sec:SMD}, the presence-absence of API calls forms a transactional dataset. In order to preserve the functionalities of original malware samples, we define   $\mathcal{S}(\mathcal{T})=\left\{\mathcal{T}'\mid \mathcal{T}'\bigcap \mathcal{T}=\mathcal{T} \right\}$. This allows us to only add some API calls to a malware sample without removing any existing ones in order to generate an adversarial example.

In our algorithm, shown as Algorithm~1, we first create a dataset of samples all identified as benign samples by the target classifier, denoted by $\mathcal{D}_\text{b}$. We next construct a code table of patterns, $CT_\text{b}$, for this dataset using the MDL principle, as described in Section~\ref{OurMDLModel}. We finally generate an adversarial example for a malware sample $\mathcal{T}$ by selecting a pattern $\mathcal{P}$ from the final code table $CT_\text{b}$, and adding it to the malware sample. The selected pattern is the one that minimizes $\ell(\mathcal{T}'\mid CT_\text{b})$ where $\mathcal{T}'=\mathcal{T}\bigcup\mathcal{P}$. 

\begin{algorithm}\label{alg:the_alg}
	\setstretch{1.1}
	\caption{MDL-based Adversarial Examples}
	\begin{algorithmic}[1]
		\State Creating $\mathcal{D}_\text{b}$ as a dataset of samples all identified as benign by the target classifier
		\State Constructing a code table of patterns, $CT_\text{b}$, using the MDL principle
		\For {each malware sample $\mathcal{T}_i$} 
		\State $\mathcal{P}^*=\underset{\mathcal{P}\in CT_\text{b}}{\arg\min}\;\ell(\mathcal{T}_i\bigcup\mathcal{P}\mid CT_\text{b})$
		\State $\mathcal{T}_{i\,\text{adv}}=\mathcal{T}_i \bigcup\mathcal{P}^*$
		\EndFor
	\end{algorithmic}	
\end{algorithm}

In our algorithm, we do not need to search among singleton patterns in $CT_\text{b}$. This is because adding a singleton item can only lead to a smaller $\ell(\mathcal{T}\bigcup\mathcal{P}\mid CT_\text{b})$ compared to the original $\ell(\mathcal{T}\mid CT_\text{b})$ if it forms a longer pattern with some of the existing items. As we check all non-singleton patterns, therefore we do not need to check singleton patterns. This is helpful as we normally have a small number of non-singleton patterns in the final code table $CT_\text{b}$, and consequently our search space is much smaller compared to considering all patterns. 

\subsection{Example}
We here provide an example for our method. In this example, we consider the dataset in the example of Section~\ref{Sec:CodeTable} as the dataset of benign samples constructed by querying the target classifier, i.e., 
\begin{align*}
\mathcal{D}&=\left\{\{1,2,3,4\}^2,\{1,2,4\},\{2,3,4,5\},\right.\\
&\hskip90pt\left.\{3,4,5\},\{4,5\},\{2\},\{3\},\{4\},\{5\}\right\}.
\end{align*}
Now let assume that we have a malware sample $\mathcal{T}_1=\left\{1,4\right\}$, and we are going to generate an adversarial example corresponding to this sample. Considering the code table for this dataset, presented in Section~\ref{Sec:CodeTable}, we can see that $\ell(\mathcal{T}_1\mid CT_\text{b})=11$, but by adding $\left\{1,2,4\right\}$ to this sample, i.e., $\mathcal{T}_1\bigcup \left\{1,2,4\right\}$, we have $\ell(\mathcal{T}_\text{1\,adv}\mid CT_\text{b})=3$. As discussed in the last section, we only need to search among non-singleton patterns which is only one pattern. This example confirms that this can make our search space much smaller. We can see that the same result is achieved by adding the singleton pattern $\{2\}$ to the original sample. This is only because this item together with the existing items $\left\{1,4\right\}$ form the longer pattern $\left\{1,2,4\right\}$. Therefore we do not need to check this pattern considering non-singleton patterns.  

\section{Performance Evaluation}
In this section, we evaluate our proposed algorithm, described in Section~\ref{Sec:proposedalg}, for the application of static malware detection in PE files where API calls are used as features.

\subsection{Dataset}
We use the dataset provided by Al-Dujaili et al.~\cite{FGSMMalware}. Our dataset consists of 14772 benign training samples, 14772 malware training samples, 4924 benign test samples, and 4924 malware test samples. The total number of API calls in the dataset is 22761. Therefore, each sample of the dataset is a binary sequence of size 22761 where the locations of ones determine API calls of that sample.

\subsection{Neural Network and Its Performance}
We use fully connected feed-forward neural networks to find the state-of-the-art performance for our dataset. We use five-fold cross validation to optimize hyper-parameters of our network. Our network consists of five layers: one input layer of size 22761, three hidden layers of size 300, and one output layer of size two. Rectified linear unit (ReLU) is used as the activation function in the hidden layers, and softmax function is used in the output layer. We use the drop out rate of 50 percent to avoid over-fitting. The size of mini-batches is 100 samples, the learning rate of Adam optimizer is 0.0001, and the number of epochs is 50. The accuracy, false positive rate (FPR), and false negative rate (FNR) obtained by this network are 91.94, 7.96, and 8.16 percent respectively. This means that the evasion rate for malware samples is 8.16 percent.

\subsection{Evasion Rate of Adversarial Examples}
We use the trained neural network presented in the last section to test our proposed algorithm for constructing adversarial examples. To construct the benign dataset $\mathcal{D}_\text{b}$ required in our algorithm, we use the benign test dataset consisting of 4924 samples, and remove the ones that are identified as malware by the trained neural network. This dataset is then used to construct the code table required for generating adversarial examples. Note that benign test samples are independent of benign training sample used to train our target neural network. 

To create a list of candidate patterns required for code table construction, we can directly use the LCM algorithm~\cite{LCM} to extract all CFPs in $\mathcal{D}_\text{b}$. However, we face pattern explosion in our dataset considering a small $minsup$. To avoid pattern explosion, we use our recently proposed approach~\cite{OurMDLClassifier} presented in Section~\ref{OurMDLModel}. We have shown that our approach acts as a pattern-summarization method by giving priority to longer patterns and without requiring to extract all CFPs. Considering our approach, we cluster $\mathcal{D}_\text{b}$ using the CLOPE algorithm~\cite{clope} with repulsion factor equal to four, and maximum cluster number equal to 16. In the CLOPE algorithm, repulsion factor is used to control intra-cluster similarity. Larger repulsion factor leads to clusters in which transactions share more common items. The clustering provides us with 16 clusters of qualities 0.20, 0.51, 0.71, 0.91, 0.75, 0.87, 0.51, 0.15, 0.50, 0.37, 0.31, 0.35, 0.34, 0.86, 0.29, and 0.08. We consider only the cluster with quality 0.08 as a low-quality cluster, and consider the remaining 15 clusters as high-quality clusters. We then apply the LCM algorithm to high-quality clusters separately with $minsup=0.001\left|\mathcal{D}_\text{b}\right|$. The list of candidate patterns is created by taking the union over the outputs of the LCM algorithm for high-quality clusters.

After creating the list of candidate patterns, we construct $CT_\text{b}$ as described in Section~\ref{Sec:CodeTable}. We finally generate one adversarial example corresponding to each malware test sample. This is done by selecting and adding a pattern from $CT_\text{b}$ to each malware test sample. The selected pattern is the one that minimizes the length of the compressed adversarial example given $CT_\text{b}$. The new evasion rate for adversarial examples is 78.24 percent which shows the effectiveness our algorithm.

\section{Discussion}
In this section, we present some discussion on the properties of our proposed method for generating adversarial examples, and also on the adversarial-training defense mechanism.

As discussed in Section~\ref{Sec:proposedmethod}, one of the main properties of our method is that we do not need to build a substitute model for the target classifier. This makes our method more practical in the scenarios where it is difficult to collect samples for all the existing classes to build a substitute model. In our method, to generate an adversarial example corresponding to a sample, we only need a dataset of samples for a wrong class. Another property of our method is that it is a general method, and can be used in different applications. Equations~\eqref{Eq: ModelSelection} and~\eqref{Eq: MDLAdversarial} presented in Section~\ref{Sec:proposedmethod} are the two key equations in our method. These equations can be made specific to a particular application. In this work, we have done this for the application of malware detection using their API calls. This is by choosing a specific family of models in the MDL principle, defining a set of constraints to preserve the functionalities of original samples, and an algorithm for finding the minimum of equation~\eqref{Eq: MDLAdversarial}.

After generating adversarial exmaples, as mentioned in the introduction, one of the main defense mechanisms is adversarial training\cite{AdversarialTraining0, AdversarialTraining1,AdversarialTraining2}. In adversarial training, both normal and adversarial examples are considered during the training process, i.e., a training dataset augmented by adversarial examples. The training dataset is augmented with both white-box and black-box adversarially generated examples. However, this method can be considered as a brute force method~\cite{AdversarialTrainingExplainability}, and has not been quite successful in improving the robustness of classifiers. As discussed by Ross and Doshi-Velez~\cite{AdversarialTrainingExplainability}, we also think that explainability/interpretability for a classifier can help us to improve its robustness in adversarial settings. We think that explainability can make adversarial training more successful by guiding us to add specific adversarial examples during the training process. Methods to interpret machine learning models are classified into two classes of intrinsic and post hoc methods~\cite{InterpretableMLBook}. Intrinsic interpretability is when a machine learning model itself is interpretable due to its structure. Post hoc interpretability is when a method is developed to interpret the decisions of a machine learning model after its training. Machine learning models that are intrinsically interpretable can also be used as a post hoc method by approximating the main model in order to explain its decisions. In our recent work, we have shown that we can use the MDL principle to build an intrinsically interpretable classifier~\cite{OurMDLClassifier}.

\section{Conclusion}
We proposed a method to generate adversarial examples using the minimum description length (MDL) principle. This is to improve the robustness of classifiers by considering these examples in their design process. We assumed that the adversary has access to only the feature set, and the final hard-decision output of the target classifier. We evaluated our method for the application of static malware detection in portable executable (PE) files. In malware detection, adversarial examples are generated by making functionality-preserving modifications to original malware samples to be misclassified as benign samples. Our method requires only a dataset of samples all identified as benign samples by the target classifier. This can be constructed by querying the target classifier. We considered a neural network to detect malware samples in PE files using their API calls. Considering API calls, a feature vector is a binary vector where the locations of ones determine existing API calls. We showed that the evasion rate is 78.24 percent for adversarial examples compared to 8.16 percent for original malware samples. This was done without changing the functionalities of malware samples. 

\bibliographystyle{IEEEtran}
%


\end{document}